\documentclass[runningheads]{llncs}
\usepackage[T1]{fontenc}
\usepackage{graphicx}
\usepackage{subcaption}
\usepackage{amsmath}
\usepackage{amssymb}
\newcommand{\norm}[1]{\left\lVert#1\right\rVert}

\usepackage{booktabs}
\usepackage{siunitx}
\usepackage{bm}

\begin{document}
\title{Explainable Deep Anomaly Detection with Sequential Hypothesis Testing for Robotic Sewer Inspection}
\titlerunning{Explainable Deep Anomaly Detection with Sequential Hypothesis Testing}
%
\author{Alex George\inst{1}\orcidID{0009-0008-5799-4161} \and
Will Shepherd\inst{2}\orcidID{0000-0003-4434-9442} \and
Simon Tait\inst{2}\orcidID{0000-0002-0004-9555} \and
Lyudmila Mihaylova\inst{1}\orcidID{0000-0001-5856-2223} \and
Sean R. Anderson\inst{1}\orcidID{0000-0002-7452-5681}} 
\authorrunning{A. George et al.}
%
\institute{School of Electrical and Electronic Engineering, University of Sheffield, Sheffield, UK\\
\email{\{ageorge4,l.s.mihaylova,s.anderson\}@sheffield.ac.uk} \and
School of Mechanical, Aerospace and Civil Engineering, University of Sheffield, Sheffield, UK\\
\email{\{w.shepherd,s.tait\}@sheffield.ac.uk}}
\maketitle              
\begin{abstract}
Sewer pipe faults, such as leaks and blockages, can lead to severe consequences including groundwater contamination, property damage, and service disruption. Traditional inspection methods rely heavily on the manual review of CCTV footage collected by mobile robots, which is inefficient and susceptible to human error. To automate this process, we propose a novel system incorporating explainable deep learning anomaly detection combined with sequential probability ratio testing (SPRT). The anomaly detector processes single image frames, providing interpretable spatial localisation of anomalies, whilst the SPRT introduces temporal evidence aggregation, enhancing robustness against noise over sequences of image frames. Experimental results demonstrate improved anomaly detection performance, highlighting the benefits of the combined spatio-temporal analysis system for reliable and robust sewer inspection.

\end{abstract}
\section{Introduction}

Sewer systems are key components of urban infrastructure, which enable the safe transportation of wastewater. However, ageing pipelines, combined with environmental and human factors, lead to the formation of cracks, leaks, blockages, root intrusions and eventually collapses~\cite{malek_2020_pipe_condition_factors}. In the UK, around 300,000 sewer blockages occur annually, costing £100 million in repairs~\cite{wateruk_blockage_report}. Traditional inspection techniques rely on manual analysis of Closed-Circuit Television (CCTV) footage collected by mobile robots, which is inefficient and prone to errors caused by fatigue and inconsistencies in interpretation by multiple operators~\cite{kumar2020_manual_inspection_issue}. The water industry is therefore investigating the potential of automated pipe condition assessment techniques that alleviate these drawbacks \cite{rayhana2021_automated_vision_research}. 

Deep learning classification algorithms have gained traction in research for sewer defect detection and have demonstrated superior performance compared to traditional machine learning methods~\cite{Haurum_2021_CVPR,li2022_vision_inspection_systems,pouyanfar2018_deeplearning_survey_advantage}. Modern explainable deep anomaly detection methods, such as Fully Convolutional Data Description (FCDD)~\cite{liznerski_explainable_2021}, can offer advantages over classification methods by predicting the spatial location of anomalies in an input image, without requiring pixel-level labelled training data. This can enhance the scene perception of autonomous mobile robots as well as aid human operators in decision support. In addition, the fully convolutional nature of FCDD offers computational efficiency compared to post-processing explainability methods like Grad-CAM \cite{selvaraju2017grad}, which necessitate additional backward passes through a classification network. Anomaly detection has been studied in sewer pipes previously using feature-based one-class classifiers such as Gabor features with Support Vector Machines and Isolation Forests \cite{Fang2020sewer}. In this paper, we propose, for the first time, the use and evaluation of an explainable deep anomaly detection system for sewer pipes based on FCDD, with feasibility evaluated on a single specific anomaly class: deposits.

A limitation of FCDD for analysing sewer CCTV  is that it operates on single image frames, disregarding the temporal dependencies in the video sequences.  This results in a potential loss of information from repeated observations of the same anomaly, as well as a potential increase in decision errors due to transient noise factors such as motion blur and variations in illumination~\cite{zhao2022_spatiotemporal_correlation}. To enhance the robustness of anomaly detection, we extend the explainable anomaly detector with the Sequential Probability Ratio Test (SPRT), a sequential hypothesis testing framework introduced by Wald~\cite{wald1992_sprt}. SPRT enables robust decision-making by aggregating anomaly scores across multiple frames. Therefore, the complete system provides spatial localisation of faults and explainability via the deep anomaly detector,  whilst the SPRT introduces
temporal processing, enhancing robustness against noise over image sequences.

The structure of this paper is as follows. In Section 2, we present the methods for deep anomaly detection and SPRT as well as the publicly available sewer pipe CCTV  datasets provided by the Water Research Centre (WRc), hosted at Spring \cite{SpringInnovation}, and from the Integrated Civil and Infrastructure Research Centre (iCAIR) at the University of Sheffield, described in \cite{edwards2023robust}. The results are then presented in Section 3, followed by a summary of the paper in Section 4.

\section{Methods}

The proposed methodology combines deep anomaly detection with sequential hypothesis testing for robust anomaly detection. An FCDD-based anomaly detector is used to generate the anomaly scores for normal and anomalous images along with an explainable anomaly heatmap. The anomaly scores are then sequentially aggregated using SPRT over time. By accumulating enough evidence, the system decides whether to classify a temporal instance as normal or anomalous. This two-stage approach ensures that the defects are detected with confidence. The following subsections cover the methodology in detail.

\subsection{Explainable Deep Anomaly Detection}

\begin{figure}
\centering
\includegraphics[scale=0.12]{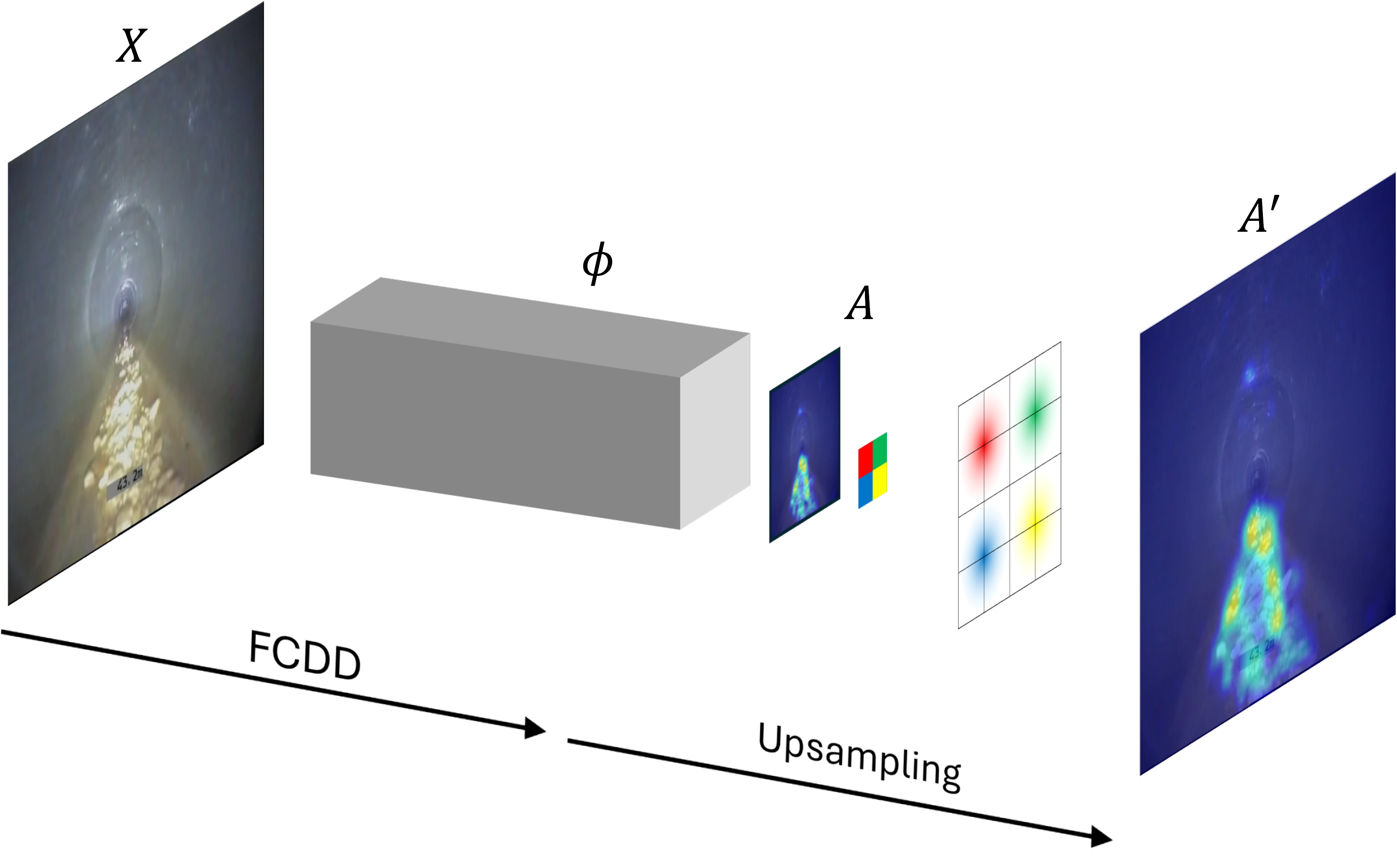}
\caption{FCDD architecture for explainable anomaly detection \cite{liznerski_explainable_2021}}
\label{fcdd_architecture}
\end{figure}

We used FCDD~\cite{liznerski_explainable_2021} to generate anomaly heatmaps $A$ and scalar anomaly scores $z$ from input images $X$ (Fig.~\ref{fcdd_architecture}), where
\begin{equation}
A\left(X\right)=\sqrt{{\phi\left(X;W\right)}^2+1}-1
\label{eq:fcdd_anomalyscore_eqn}
\end{equation}
and  $\phi$ is a fully convolutional network (FCN) with weights $W$. The anomaly heatmap  $A$ is upsampled using a strided transposed convolution with a Gaussian kernel to generate $A'$, which has direct pixel-wise correspondence to $X$. The anomaly score $z$  is obtained from the L1-norm of $A$,
\begin{equation}
z = \frac{1}{u\cdot v}\norm{A}_{1} 
\label{eq:mean_anomaly_score}
\end{equation}
where $u$ and $v$ correspond to the dimensions of $A$. 

FCDD was trained using the following loss function, 
\begin{equation}
\min_{W} \ \frac{1}{n}\sum^{n}_{i=1}(1-y_i)\frac{1}{u\cdot v}\norm{A(X_i)}_{1}-y_i\log\left(1-\exp\left(-\frac{1}{u\cdot v}\norm{A(X_i)}_{1}\right)\right)
\label{eq:fcdd_objective_fn}
\end{equation}
where $y_i=0$ indicates a normal sample and $y_i=1$ indicates an anomaly. The first three stages of a pre-trained Inception-v3 network were used as a backbone for FCDD. Training was performed using the Adam optimiser with a learning rate of $10^{-4}$ and a mini-batch size of 32. Data augmentation techniques (rotation, translation, jittering and noise) were applied randomly to 50\% of the training images. Anomaly score predictions using a calibration dataset were used to tune a threshold parameter $\tau$ using ROC curve analysis to determine normal $(z<\tau)$ and anomalous samples $(z\geq\tau)$.

\subsection{Sequential Probability Ratio Test}

We integrate time-dependent processing into anomaly detection using the Sequential Probability Ratio Test (SPRT) framework~\cite{wald1992_sprt}. SPRT is a form of sequential hypothesis testing that accumulates evidence from anomaly scores over time to determine if a normal ($H_0$) or anomalous ($H_1$) state is supported. Let $z_1, z_2,...,z_t$ represent the anomaly scores from the detector for video image frames up to time $t$. The cumulative sum of the log-likelihood ratio at time $t$ is then calculated as
\begin{equation}
\Lambda_t = \Lambda_{t-1} + \log \frac{p(z_t | H_1)}{p(z_t | H_0)}
\label{eq:sprt_eqn}
\end{equation}
where $p(z_t | H_1)$ and $p(z_t | H_0)$ are the probability density functions (PDFs) of $z_t$ under each hypothesis. In this case, $p(z_t | H_1)$ was modelled as a Gaussian mixture model and $p(z_t | H_0)$ was modelled as a gamma distribution.

The decision rule in SPRT is 
\begin{align}
\Lambda_t\leq a: & \quad \text{accept } H_0 \text{ and reject } H_1 \text{ (Normal)} \\
\Lambda_t\geq b: & \quad \text{accept } H_1 \text{ and reject } H_0 \text{ (Anomaly)} \\
a< \Lambda_t<b: & \quad \text{continue monitoring (Undecided)}
\end{align}
where $a$ and $b$ are the stopping bounds for accepting and rejecting a hypothesis. Note that  $\Lambda_t$ is reinitialised to zero whenever a decision boundary is crossed.

The values for $a$ and $b$ are derived from the user-specified Type I error probability ($\alpha$, or false positive rate) and Type II error probability ($\beta$, or false negative rate), respectively, and are approximated as
\begin{equation}
a \approx \log \frac{\beta}{1-\alpha} \text{ and } b \approx \log \frac{1-\beta}{\alpha}.
\label{eq:type1type2_eqn}
\end{equation}
Parameter selection for $\alpha$ and $\beta$ involves a trade-off: $\alpha < \beta$  minimises false positives at the cost of missed faults, while $\alpha > \beta$  prioritises fault detection, whilst potentially increasing false positives. In this investigation, we used $\alpha < \beta$ to reduce false positives.


\subsection{Experimental Data}

The FCDD network was trained and evaluated using two distinct publicly available datasets: the single image frame WRc dataset \cite{SpringInnovation} and the iCAIR video dataset \cite{edwards2023robust}. For the WRc dataset, FCDD was trained in a one-vs-all framework with 2500 anomaly (deposit) and 5005 normal (non-deposit) image frames. The model was calibrated on separate data (1001 anomaly, 1012 normal) and evaluated on a further independent test set (2001 anomaly, 2006 normal). For benchmarking, two binary Support Vector Machine (SVM) classifiers were also trained and evaluated on the WRc data, one employing Gabor features and the other utilising features extracted from a pre-trained Inception-v3 model. The iCAIR video dataset, featuring deposits as anomalies, was used to train (25 anomaly frames, 400 normal frames), calibrate (75 anomaly frames, 225 normal frames), and evaluate (5475 frames) FCDD and FCDD plus SPRT.


\begin{figure}
\label{fig:comparison}
    \centering
    \begin{minipage}{0.48\textwidth}
        \centering
        \begin{subfigure}{\textwidth}
            \includegraphics[width=\linewidth]{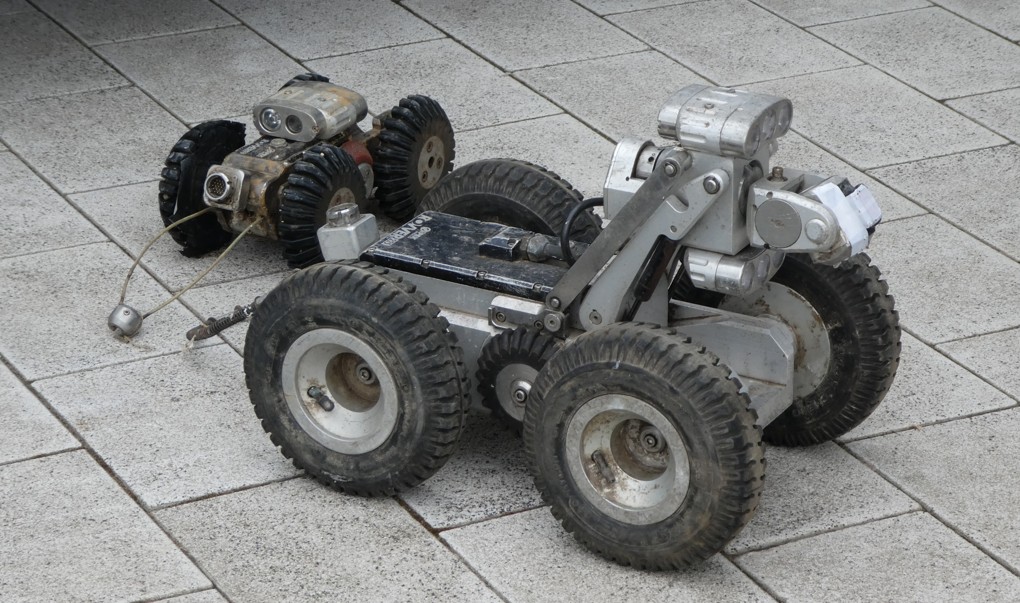}
            \subcaption{}
        \end{subfigure}
        \\[1.5mm] 
        \begin{subfigure}{0.48\textwidth}
            \includegraphics[width=\linewidth]{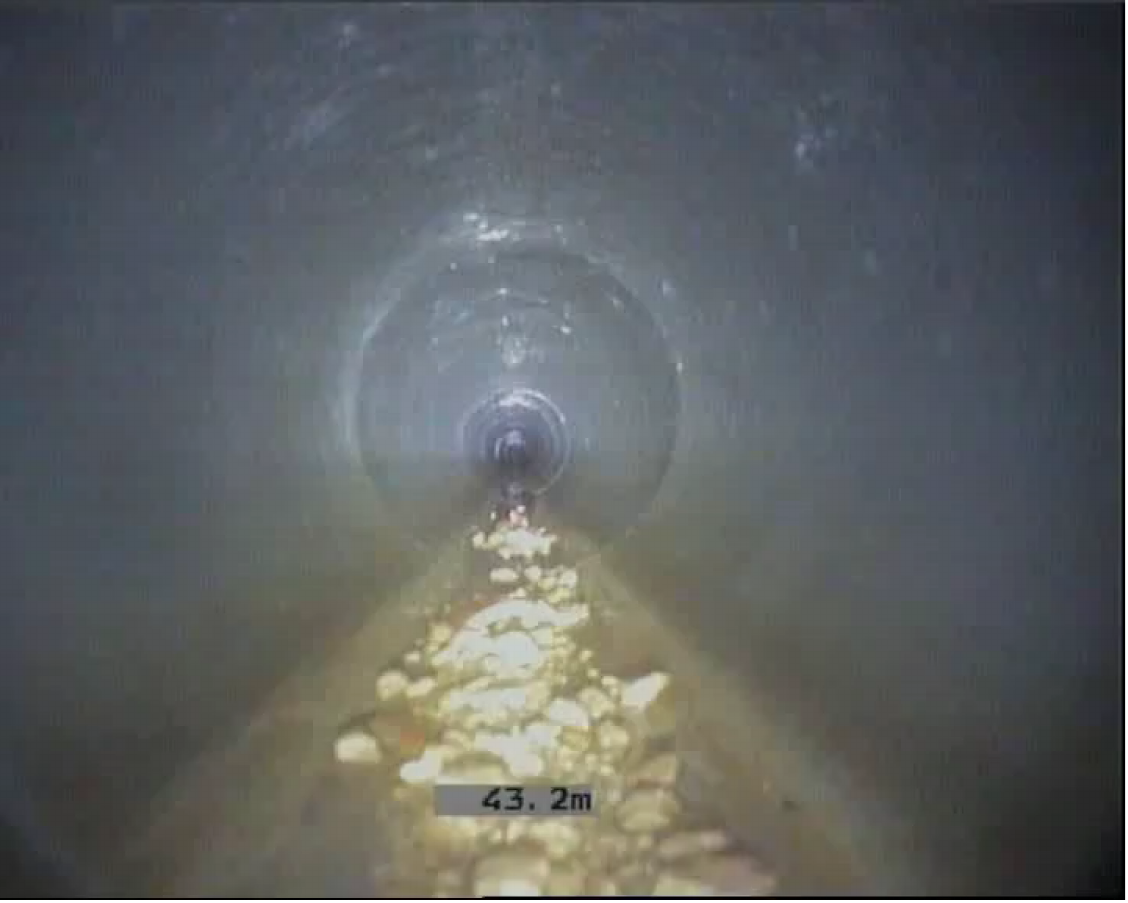}
            \subcaption{}
        \end{subfigure}
        \begin{subfigure}{0.48\textwidth}
            \includegraphics[width=\linewidth]{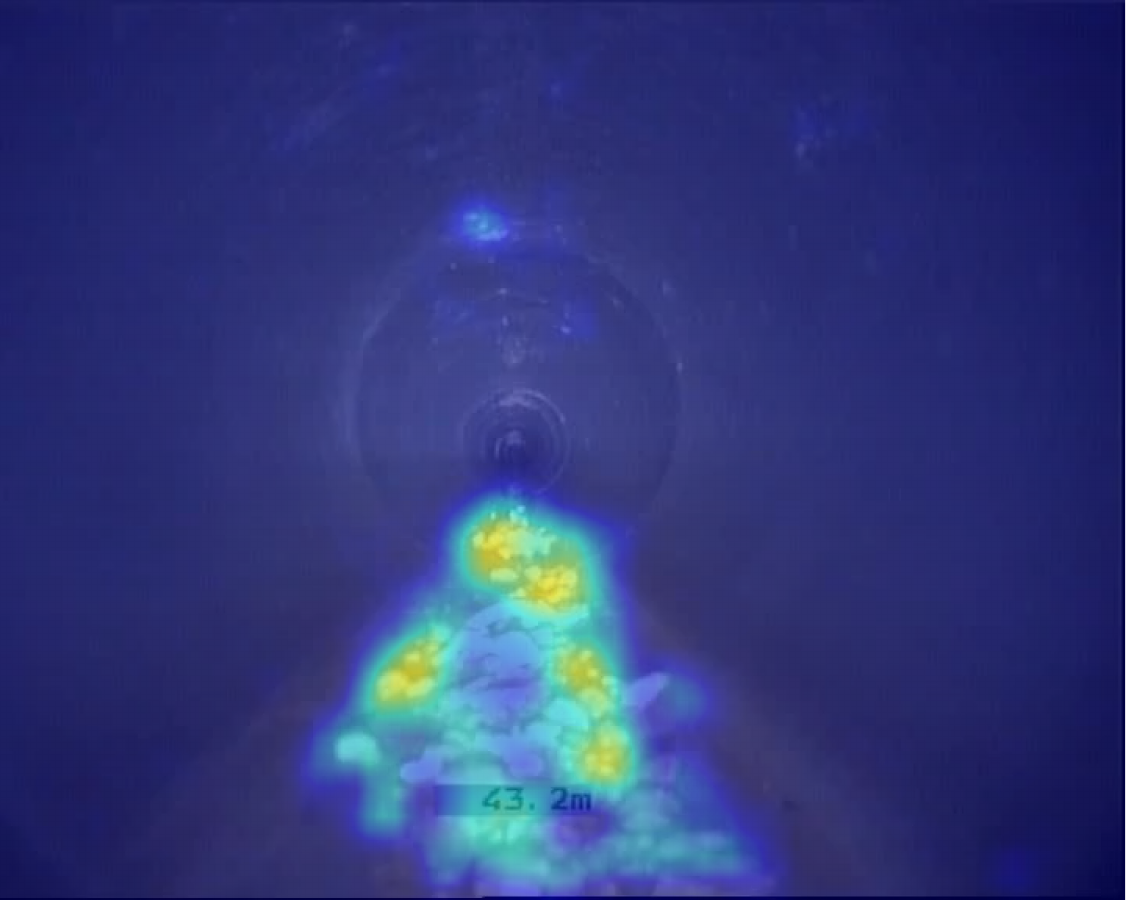}
            \subcaption{}
        \end{subfigure}
    \end{minipage}
    \begin{minipage}{0.48\textwidth}
        \centering
        \begin{subfigure}{\textwidth}
            \includegraphics[width=\linewidth]{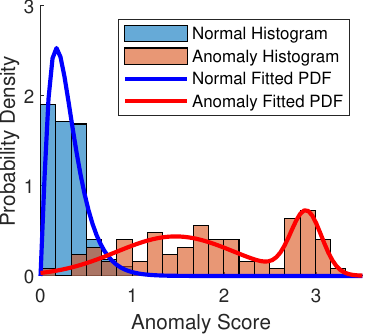}
            \subcaption{}
        \end{subfigure}
    \end{minipage}
    \caption{(a) Sewer inspection robot. (b) Example of CCTV image with anomaly. (c) CCTV image with anomaly heatmap overlay predicted by FCDD. (d) Anomaly score distributions from the trained FCDD system, with fitted PDFs for normal $p(z_t | H_0)$ and anomalous distributions $p(z_t | H_1)$.}
\end{figure}

\vspace{-1em}

\section{Results}

In sewer CCTV inspection, a robot is typically used to capture image frames.  Fig. 2(a)-(c) shows the robot from the iCAIR study along with an example input image frame $X$ captured by CCTV and an example of the anomaly heatmap predicted by FCDD, overlaid on the input image. Note that the spatial location of the deposit data is predicted in Fig. 2(c) by FCDD, even though it is not explicitly trained to do this by pixel-wise labelling. A histogram of anomaly scores predicted by FCDD on the iCAIR calibration dataset for both normal and anomaly data are shown in Fig. 2(d). 

To evaluate FCDD on single image frames, we used the large WRc dataset, comparing SVM models trained on Gabor features and Inception-v3 features. We chose the optimal threshold for FCDD to be the maximum of Youden's index, i.e. the point on the ROC curve (for the calibration data) with the maximum difference between the true positive rate (TPR) and false positive rate (FPR). Performance was substantially better using FCDD, with F1-scores of 88.03\% using FCDD vs 44.94\% and 71.72\% using the SVM models (Table \ref{tab:results}).

\begin{figure}[htp]
\centering
\includegraphics[scale=0.9]{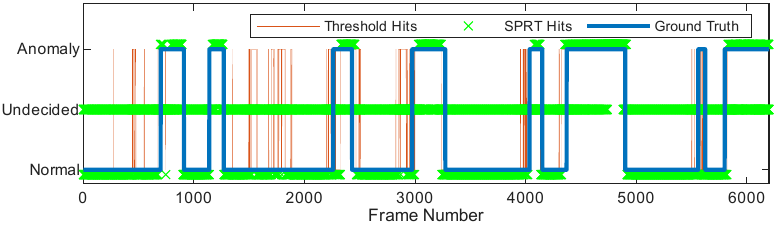}
\caption{Anomaly predictions  for FCDD on the iCAIR video frames from thresholding ($\tau\mathbin{=}0.64$, red) and SPRT ($\alpha\mathbin{=}10^{-6}, \beta\mathbin{=}0.01$, green) overlaid on ground truth (blue). } 
\label{score_plot}
\end{figure}

Quantitative analysis on the smaller iCAIR video dataset on a per-frame basis demonstrated improved F1-scores when combining FCDD with SPRT (94.36\%) when $\alpha=10^{-6}$ and $\beta=0.01$ (chosen to demonstrate the effect of reducing false positives whilst maintaining detections), compared to simple thresholding (84.07\% - Table \ref{tab:results}). However, it should also be noted that the total frame count used in SPRT's metric calculations was reduced due to undecided frames between threshold boundaries, thereby focusing the metrics on the quality of definitive decisions. Fig.~\ref{score_plot} demonstrates that SPRT reduces false alarms associated with single-frame thresholding, although misses one anomaly at approximately frame 5600 due to lack of evidence. This illustrates the fundamental trade-off between the choice of values for $\alpha$ and $\beta$ to provide an optimal balance between decision robustness gained through evidence accumulation and reduced sensitivity to fleeting or less persistent anomalies. The model and corresponding code are made publicly available on GitHub \cite{alex_github}.

\vspace{-2em}

\begin{table}[htp]
\centering
\caption{Model performance comparison with WRc and iCAIR datasets}
\label{tab:results}
\begin{tabular}{llccccc}
\toprule
Dataset & Method & {Accuracy} & {Precision} & {Recall (TPR)} & {FPR} & {F1-Score} \\
\midrule
WRc & SVM (Gabor) & 62.94 & 87.07 & 30.28 & \textbf{4.49} & 44.94 \\
(single & SVM (Inception-v3) & 76.24 & 88.42 & 60.32 & 7.88 & 71.72 \\
images)& FCDD $(\tau=1.09)$ & \textbf{88.30} & \textbf{89.98} & \textbf{86.16} & 9.57 & \textbf{88.03} \\
\midrule
iCAIR & FCDD $(\tau=0.64)$ & 88.79 & 79.69 & 88.96 & 11.30 & 84.07 \\
(video) & FCDD+SPRT  & \textbf{96.47} & \textbf{97.85} & \textbf{91.12} & \textbf{0.96} & \textbf{94.36} \\
\bottomrule
\end{tabular}
\end{table}

\vspace{-3em}
  
\vspace{-2em}

\section{Summary}

This paper has proposed a novel sewer pipe fault detection system using explainable deep anomaly detection, which can predict the spatial locations of anomalies. The deep anomaly detector was also combined with sequential hypothesis testing via the SPRT. Experimental results demonstrated that FCDD improved anomaly detection over an SVM method, and that SPRT enhanced the robustness of anomaly detection by aggregating temporal information. 

\begin{credits}
\subsubsection{\ackname} This work is supported by the European Union’s Horizon Europe Research and Innovation Programme under Grant Agreement No. 101189847 - Pipeon.

\subsubsection{\discintname}

The authors have no competing interests to declare that are
relevant to the content of this article.

\end{credits}
%
%
%
\bibliographystyle{splncs04}
\bibliography{bibliography}

\end{document}